\documentclass[lettersize,journal]{IEEEtran}
\usepackage{amsmath,amsfonts,amssymb}
\usepackage{algorithmic}
\usepackage{algorithm}
\usepackage{array}
\usepackage{textcomp}
\usepackage{stfloats}
\usepackage{url}
\usepackage{verbatim}
\usepackage{graphicx}
\usepackage{cite}
\usepackage{multicol,multirow}
\usepackage{subfigure}

\begin{document}

\title{A Temporally Disentangled Contrastive Diffusion Model for Spatiotemporal Imputation}

\author{Yakun Chen,~\IEEEmembership{Student Member,~IEEE}, Kaize Shi,~\IEEEmembership{Member,~IEEE}, Zhangkai Wu,~\IEEEmembership{Student Member,~IEEE}, Juan Chen, Xianzhi Wang,~\IEEEmembership{Member,~IEEE}, Julian McAuley, Guandong Xu,~\IEEEmembership{Member,~IEEE}, Shui Yu,~\IEEEmembership{Fellow,~IEEE}

\thanks{Y. Chen, Z. Wu, K. Shi, X. Wang, G. Xu, and S. Yu are with School of Computer Science, University of Technology Sydney, Australia.}
\thanks{J. Chen is with School of Computer Science and Technology, Zhejiang Normal University, China.}
\thanks{J. McAuley is with Department of Computer Science and Engineering, University of California, San Diego, USA.}
}

\markboth{Journal of \LaTeX\ Class Files,~Vol.~14, No.~8, August~2021}%
{Shell \MakeLowercase{\textit{et al.}}: A Sample Article Using IEEEtran.cls for IEEE Journals}


\maketitle

\begin{abstract}
Spatiotemporal data analysis is pivotal across various domains, such as transportation, meteorology, and healthcare. The data collected in real-world scenarios are often incomplete due to device malfunctions and network errors. Spatiotemporal imputation aims to predict missing values by exploiting the spatial and temporal dependencies in the observed data. Traditional imputation approaches based on statistical and machine learning techniques require the data to conform to their distributional assumptions, while graph and recurrent neural networks are prone to error accumulation problems due to their recurrent structures.
Generative models, especially diffusion models, can potentially circumvent the reliance on inaccurate, previously imputed values for future predictions; 
However, diffusion models still face challenges in generating stable results.
We propose to address these challenges by designing conditional information to guide the generative process and expedite the training process.
We introduce a conditional diffusion framework called C$^2$TSD, which incorporates disentangled temporal (trend and seasonality) representations as conditional information and employs contrastive learning to improve generalizability.
Our extensive experiments on three real-world datasets demonstrate the superior performance of our approach compared to a number of state-of-the-art baselines.
\end{abstract} 


\section{Introduction}
Spatiotemporal data represent the type of multivariate time series data concerning both location and time.
They typically differ from other data types in being chronologically ordered and autocorrelated---they may exhibit trends, seasonality, and noise while showing non-stationarity overall~\cite{wu2019graph}.
Spatiotemporal data widely serve as valuable assets to generate insights and support decision-making in economics, transportation, healthcare, meteorology, etc~\cite{li2020multimodal,tedjopurnomo2020survey,di2023explainable,bauer2015quiet}.
While existing techniques that deal with spatiotemporal data (e.g., forecasting~\cite{han2019review}, classification~\cite{ismail2019deep}, and anomaly detection~\cite{blazquez2021review}) generally rely on a complete spatiotemporal dataset to uphold their performance, spatiotemporal data inevitably contain missing values due to sensor malfunctions, packet loss in data transmission, poor management of data integrity, or human neglect in practical scenarios~\cite{che2018recurrent}. Those missing values could significantly degrade the performance of specific tasks, calling for effective imputation methods to make up the missing values to facilitate downstream tasks.

Spatiotemporal imputation generally requires capturing spatial and temporal dependencies simultaneously~\cite{cini2021filling}.
Traditional statistical and machine-learning-based imputation methods require labor-intensive feature engineering and typically make strong assumptions about data, which may not always hold in real-world scenarios.
Although Recurrent Neural Network (RNN)-based models, including those based on Gated Recurrent Units (GRUs)~\cite{che2018recurrent}, bidirectional structures~\cite{cao2018brits}, and the integration of Graph Neural Networks (GNNs) and RNNs~\cite{cini2021filling}, have shown advantages in capturing complicated relations,
they suffer the error accumulation issue due to their recurrent structures~\cite{wang2024deep,liu2019naomi}, i.e., they heavily rely on previously imputed (and potentially inaccurate) values to infer more missing values, leading to degraded imputation performance. 

\begin{figure}[!t]
\centering
\includegraphics[width=0.42\textwidth]{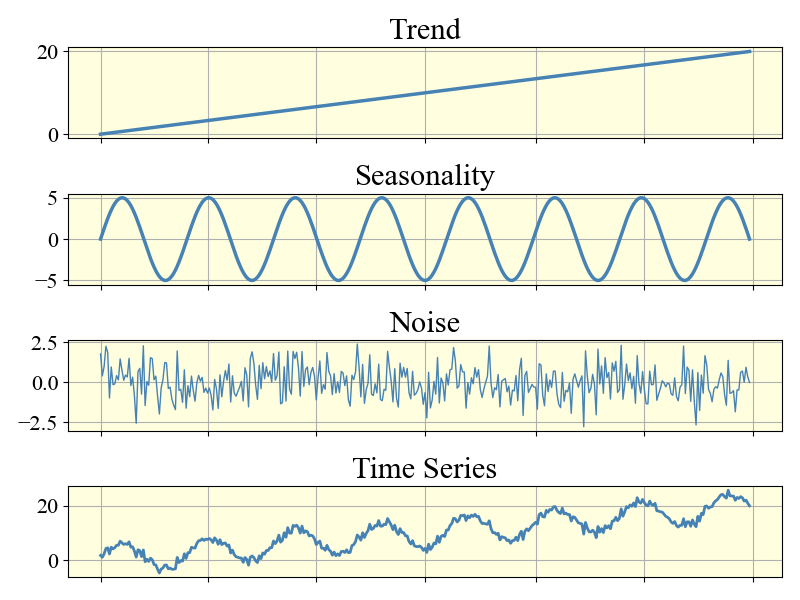}
\caption{An illustrative example of trend, seasonality, and noise components (the top three boxes) that constitute a time series (at the bottom).
} 
\label{fig:disentangle}
\end{figure}


Diffusion models offer a solution to the error accumulation problem---they gradually add random noise to data and learn to remove the noise in a reverse process while estimating missing values~\cite{lin2023diffusion}.
As such, diffusion models can generate probabilistic results that are more suitable for dealing with non-stationary spatiotemporal data.
Recent efforts~\cite{tashiro2021csdi,liu2023pristi} have adapted Denoising Diffusion Probabilistic Models (DDPMs)~\cite{ho2020denoising} for spatiotemporal imputation.
They introduce conditional information as a guide in the reverse process to improve the predictability of DDPMs' final output.
However, using the observed data directly as conditional information, they capture time series components (trend, seasonality, noise) as entangled temporal dynamics; thus, the imputation results are impacted by noise inherent to observed data.

We aim to address to above limitations by combining the benefits of diffusion models and structural time series models. 
The principles of structural time series models~\cite{qiu2018multivariate} are to disentangle time series as a combination of trend, seasonality, and noise components (illustrated in Figure~\ref{fig:disentangle}).
While the noise is generally considered unpredictable, the disentanglement of trend and seasonality enables the extraction of separate temporal representations of the observed data, enhancing the model's ability to learn complex temporal dependencies~\cite{wen2019robuststl}.
Such disentanglement has proven effective in time series forecasting, especially when enhanced with contrastive learning~\cite{woo2022cost}, showing great potential for spatiotemporal imputation.

Inspired by the above, we propose C$^2$TSD, a \textbf{\underline C}onditional diffusion framework integrating \textbf{\underline C}ontrastive learning and \textbf{\underline T}rend-\textbf{\underline S}eason \textbf{\underline D}isentanglement for spatiotemporal imputation.
Specifically, we use a conditional diffusion model as the backbone to generate probabilistic results to handle non-stationary spatiotemporal data and mitigate the error accumulation problem.
The design of conditional information helps guide the generation of high-quality data that conforms to preset conditions.
To incorporate spatiotemporal dependencies in the conditional information, we first use trend and seasonal extraction to learn the disentangled temporal dependencies, followed by a GNN encoder to aggregate the geographical neighbor information to learn the spatial dependencies. 
Specifically, C$^2$TSD employs temporal and spatial attention modules as the noise prediction model to denoise and reconstruct the data distribution.
C$^2$TSD also applies a Contrastive learning strategy to improve the generalization ability on unfamiliar and unseen sample distributions.

Our contributions are summarized as follows:
\begin{itemize}
    \item We propose C$^2$TSD, a contrastive diffusion framework for spatiotemporal imputation, which constructs and uses conditional information with disentangled temporal representations and spatial relationships.
    \item We introduce trend-season disentanglement to enhance the robustness of temporal relationship learning. The disentanglement enables independently discerning trend and seasonal dependencies, allowing our framework to handle non-stationary spatiotemporal data effectively.
    \item We develop a contrastive learning strategy to aid the learning of spatiotemporal dependencies to enhance the stability and generalization of our proposed model. 
    \item We have conducted extensive experiments on three real-world datasets from meteorology and transportation fields, which demonstrate the superior performance of C$^2$TSD in spatiotemporal imputation compared to multiple state-of-the-art baselines.
\end{itemize}

The remainder of this paper is structured as follows.
Section~\ref{sec:method} introduces our proposed model, including its pipeline and major components.
Section~\ref{sec:experiment} reports our empirical evaluation. 
Section~\ref{sec:related} reviews the related work.
And finally, Section~\ref{sec:conclusion} gives the concluding remarks.

\begin{figure*}[!t]
\centering
\includegraphics[width=\textwidth]{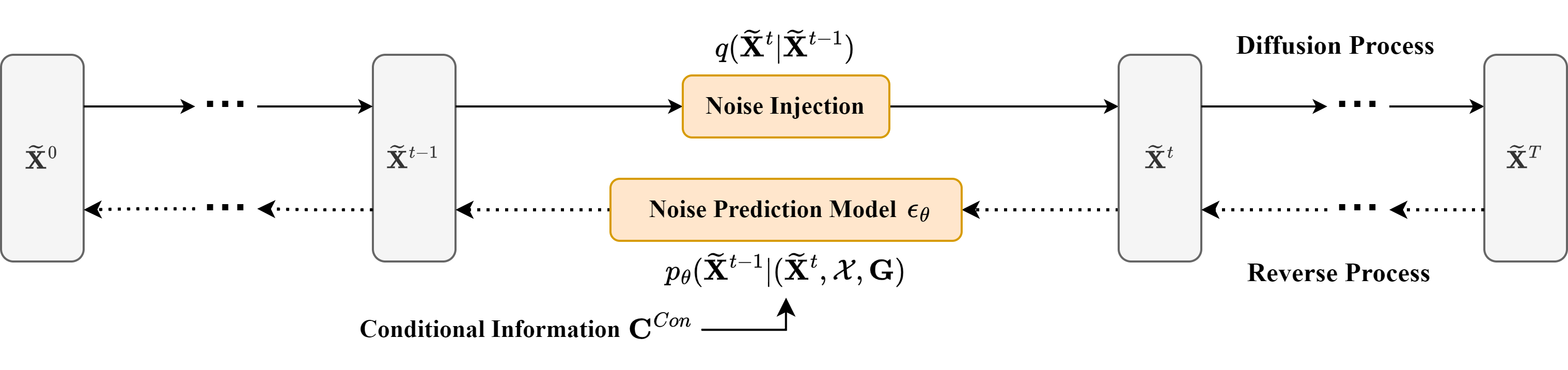}
\caption{
Architecture of \textbf{C$^2$TSD}. Our framework follows the pipeline of denoising diffusion probabilistic models. In particular, C$^2$TSD uses a trained noise prediction model $\epsilon_\theta$ to sample $\widetilde{\mathbf{X}}^{t-1}$ step by step in the reverse process, under the guidance of the conditional information $\mathbf{C}^{\mathit{Con}}$ generated from interpolated observed data $\mathcal{X}$ and geographical information $\mathbf{G}$. It also uses a contrastive loss to supplement the reconstruction loss of the diffusion model.
} 
\label{fig:framework}
\end{figure*}




\section{Approach}
\label{sec:method}
In this section, we present the problem definition and our proposed approach for spatiotemporal imputation, followed by introducing its key components.

\subsection{Problem Statement}
Spatiotemporal imputation aims to predict missing values in historical records from both spatial and temporal dimensions. We formulate the input spatiotemporal data as $\mathbf{X}_{1:L} = \{ \mathbf{X}_{1},\ldots, \mathbf{X}_{l},\ldots, \mathbf{X}_{L} \} \in \mathbb{R}^{N \times L}$, where $\mathbf{X}_{l}\in \mathbb{R}^{N}$ 
are observed values at time $l$ for $N$ nodes, e.g., air quality monitoring stations and traffic speed sensor stations.
Unlike forecasting tasks, all timestamps are available for training and testing in imputation tasks. A mask matrix $\mathbf{M} \in {\{0,1\}}^{N \times L}$ has the same dimensions as the input; it identifies the locations of missing values in spatiotemporal data, where $m^{n}_{l}=0$ denotes $x^{n}_{l}$ is missing in data collection, and $m^{n}_{l}=1$ otherwise. The output $\widehat{\mathbf{Y}}_{1:L} \in \mathbb{R}^{N \times L}$ matches the input dimensions and is expected to contain predicted values for the missing location in $\mathbf{M}$. Thus, spatiotemporal imputation aims to predict values most closely approximate the underlying ground truth, filling the missing gaps in $\mathbf{X}_{1:L}$, thereby generating a complete dataset for future downstream tasks.

\subsection{Approach Overview}
Our approach (Fig.~\ref{fig:framework}) is fundamentally based on denoising diffusion probabilistic models (DDPMs)~\cite{ho2020denoising}.
Since diffusion models pose no limitation on the architecture of the noise prediction model, they offer the flexibility to customize the design of the noise prediction model to the requirements of specific tasks.
Leveraging this edge, C$^2$TSD differs from DDPMs in deriving conditional information from the interpolated observed data and geographical information to guide the noise prediction model in learning the spatiotemporal dependencies.
We also incorporate contrastive learning to facilitate the learning of discriminative features to generalize the model.


Specifically, C$^2$TSD generates samples consistent with the original data distribution by systematically introducing noise into the samples and subsequently learning the reverse denoising process. Similar to a generic diffusion model, C$^2$TSD can be identified as two Markov Chain processes with $T$ steps: a \textit{diffusion process} and a \textit{reverse process}. Starting with $\widetilde{\mathbf{X}}^0 \sim p_{\mathit{data}}$, where $p_{\mathit{data}}$ denotes the original data distribution, $\widetilde{\mathbf{X}}^t$ represents the sampled latent variable sequence for diffusion steps $t = 1, \ldots, T$. At the final step, $\widetilde{\mathbf{X}}^T$ approximates a Gaussian distribution, $\mathcal{N}(0,\mathbf{I})$. The diffusion process gradually injects Gaussian noise into $\widetilde{\mathbf{X}}^0$ until close to $\widetilde{\mathbf{X}}^T$, whereas the reverse process aims to denoise $\widetilde{\mathbf{X}}^t$ to revert it to $\widetilde{\mathbf{X}}^0$.

We consider the spatiotemporal imputation task as a conditional generation task to apply diffusion models~\cite{tashiro2021csdi,liu2023pristi}.
We aim to estimate the conditional probability distribution $q(\widetilde{\mathbf{X}}^0_{1:L}|\mathbf{X}_{1:L})$, where the imputation of $\widetilde{\mathbf{X}}^0_{1:L}$ is conditioned by the observed values $\mathbf{X}_{1:L}$.
We use the superscript $t \in \{ 0,1,\ldots, T \}$ to denote the diffusion steps and omit the subscripts $1:L$ (time length for spatiotemporal data) for brevity in the following discussion.

The \textit{diffusion process} for spatiotemporal imputation injects Gaussian noise step by step into the input data within the imputation target without considering the conditional information, which can be formulated as:  
\begin{equation}
    \begin{split}
        q(\widetilde{\mathbf{X}}^{1:T}|\widetilde{\mathbf{X}}^0) &= \prod_{t=1}^{T} q(\widetilde{\mathbf{X}}^t|\widetilde{\mathbf{X}}^{t-1}), \\
        q(\widetilde{\mathbf{X}}^t|\widetilde{\mathbf{X}}^{t-1}) &= \mathcal{N}(\widetilde{\mathbf{X}}^t; \sqrt{1-\beta_t}\widetilde{\mathbf{X}}^{t-1}, \beta_t \mathbf{I}),       
    \end{split}
    \label{eq1}
\end{equation}
where $\beta_t \in (0,1)$ is a variance schedule of the diffusion process and is always chosen ahead of model training. $\widetilde{\mathbf{X}}^t$ is derived through the sampling process $\widetilde{\mathbf{X}}^t = \sqrt{\overline{\alpha}_t}\widetilde{\mathbf{X}}^0 + \sqrt{1-\overline{\alpha}_t}\boldsymbol{\epsilon}$, where $\alpha_t = 1-\beta_t$, $\overline{\alpha}_t=\prod_{i=1}^{t}\alpha_i$, and $\epsilon$ represents the sampled standard Gaussian noise. As $T$ increases sufficiently, the distribution $q(\widetilde{\mathbf{X}}^T|\widetilde{\mathbf{X}}^0)$ approximates a standard normal distribution.

The \textit{reverse process} gradually converts random noise to imputed values based on conditional information. Following~\cite{liu2023pristi}, the reverse process is conditioned on the interpolated conditional information $\mathcal{X}$ to enhance the observed values and geographical information $\mathbf{G}$, which can be formulated as:
\begin{equation}
    \begin{split}
        p_{\theta}(\widetilde{\mathbf{X}}^{0:T-1}|\widetilde{\mathbf{X}}^T, \mathcal{X}, \mathbf{G}) &= \prod_{t=1}^{T} p_{\theta}((\widetilde{\mathbf{X}}^{t-1}|(\widetilde{\mathbf{X}}^t, \mathcal{X}, \mathbf{G}), \\
        p_{\theta}(\widetilde{\mathbf{X}}^{t-1}|(\widetilde{\mathbf{X}}^t, \mathcal{X}, \mathbf{G}) &= \mathcal{N}((\widetilde{\mathbf{X}}^{t-1}; \mu_{\theta}((\widetilde{\mathbf{X}}^t, \mathcal{X}, \mathbf{G}, t), \sigma^2_{\theta}\mathbf{I}).
    \end{split}
\end{equation}

Ho et al.~\cite{ho2020denoising} introduce an effective parameterization of $\mu_\theta$ and $\sigma^2_{\theta}$, which can be formulated as:
\begin{equation}
    \begin{split}
        \mu_{\theta}(\widetilde{\mathbf{X}}^{t}, \mathcal{X}, \mathbf{G}, t) &= \frac{1}{\sqrt{\overline{\alpha_t}}}\left(\widetilde{\mathbf{X}}^{t} - \frac{\beta_t}{\sqrt{1-\overline{\alpha_t}}} \epsilon_{\theta}(\widetilde{\mathbf{X}}^{t}, \mathcal{X}, \mathbf{G}, t)\right), \\
        \sigma^2_t &= \frac{1-\overline{\alpha}_{t-1}}{1-\overline{\alpha}_t}\beta_t,
    \end{split}
\label{eq:reverse}
\end{equation}
where $\epsilon_\theta$ represents a trainable denoising neural network parameterized by $\theta$. It accepts the noisy sample $\widetilde{\mathbf{X}}^t$, conditional information $\mathcal{X}$, and geographical information $\mathbf{G}$ as inputs to predict the added noise $\epsilon$ on imputation target to reconstruct the original information within the noisy sample. $\epsilon_\theta$ is commonly called \textit{noise prediction model}.

\subsubsection{Training Process}

During the training process (Algorithm~\ref{alg:training}), we randomly mask some positions of the input data $\mathbf{X}$ to generate the imputation target $\widetilde{\mathbf{X}}^t$.
The remaining observations serve as conditional information for the imputation process.
Following~\cite{tashiro2021csdi}, we apply point and block mask strategies to generate imputation targets with different missing patterns.
With the generated imputation targets $\widetilde{\mathbf{X}}^0$ and the interpolated conditional information $\mathcal{X}$, the training objective for spatiotemporal imputation can be formulated as:
\begin{equation}
    \mathcal{L}_{rl}(\theta) = \mathbb{E}_{\widetilde{\mathbf{X}}^0 \sim q_{\theta}(\widetilde{\mathbf{X}}^0), \epsilon \sim \mathcal{N}(0, \mathbf{I})} \left\lVert \epsilon - \epsilon_{\theta}(\widetilde{\mathbf{X}}^t, \mathcal{X}, \mathbf{G}, t) \right\rVert^2.
\label{eq:losstheta}
\end{equation}

During each iteration of the training process, we sample the Gaussian noise $\epsilon$, the imputation target $\widetilde{\mathbf{X}}^t$, the diffusion step $t$, and calculate the interpolated conditional information $\mathcal{X}$ based on the remaining observations. 

\begin{algorithm}[t]
\caption{Training process of C$^2$TSD}
\label{alg:training}
\begin{algorithmic}[1]
\STATE \textbf{Input}: Incomplete observed data $\mathbf{X}$, the adjacency matrix $\mathbf{A}$, the number of iteration $N_{it}$, the number of diffusion steps $T$, noise levels sequence $\overline{\alpha}_t$.
\STATE \textbf{Output}: Optimized noise prediction model $\epsilon_\theta$.
\FOR{$i = 1$ \TO $N_{it}$}{
    \STATE $\widetilde{\mathbf{X}}^0 \leftarrow \text{Mask}(\mathbf{X})$;
    \STATE $\mathcal{X} \leftarrow \text{Interpolate}(\widetilde{\mathbf{X}}^0)$;
    \STATE Sample $t \sim \text{Uniform}(\{1, \ldots, T\}), \epsilon \sim \mathcal{N}(0, \mathbf{I})$;
    \STATE $\widetilde{\mathbf{X}}^t \leftarrow \sqrt{\overline{\alpha}_t}\widetilde{\mathbf{X}}^0 + \sqrt{1-\overline{\alpha}_t}\epsilon$;
    \STATE Update the gradient $\nabla_{\theta} \lVert \epsilon - \epsilon_{\theta}(\widetilde{\mathbf{X}}^t, \mathcal{X}, \mathbf{G}, t) \rVert^2$.
}\ENDFOR
\end{algorithmic}
\end{algorithm}

\subsubsection{Imputation Process}
During the imputation process (Algorithm~\ref{alg:imputation}), we use the noise prediction model $\epsilon_\theta$ trained during the training process. The observed mask $\widetilde{M}$ is available, and the imputation target $\widetilde{X}$ in the current process is all missing values within the spatiotemporal dataset.
After calculating the interpolated conditional information $\mathcal{X}$ based on all observed values, the model takes $\widetilde{\mathbf{X}}^t$ and $\mathcal{X}$ as inputs to generate samples of imputation results according to Eq.~\eqref{eq:reverse}.


\begin{algorithm}[t]
\caption{Imputation process with C$^2$TSD}
\label{alg:imputation}
\begin{algorithmic}[1]
\STATE \textbf{Input:} A sample of incomplete observed data $\mathbf{X}$, the adjacency matrix $A$, the number of diffusion steps $T$, the optimized noise prediction model $\epsilon_\theta$.
\STATE \textbf{Output:} Missing values of the imputation target $\widehat{X}^0$.
\STATE $\mathcal{X} \gets \text{Interpolate}(\mathbf{X})$;
\STATE Set $\widetilde{\mathbf{X}}^T \sim \mathcal{N}(0,I)$;
\FOR{$t = T$ \TO $1$}{
    \STATE $\mu_{\theta}(\widetilde{\mathbf{X}}^{t}, \mathcal{X}, \mathbf{G}, t) \leftarrow \frac{1}{\sqrt{\overline{\alpha}_t}} \left( \widetilde{\mathbf{X}}^t - \frac{\beta_t}{\sqrt{1-\overline{\alpha}_t}} \epsilon_{\theta}(\widetilde{\mathbf{X}}^{t}, \mathcal{X}, \mathbf{G}, t) \right)$;
    \STATE $\widetilde{\mathbf{X}}^{t-1} \leftarrow \mathcal{N}(\mu_{\theta}(\widetilde{\mathbf{X}}^{t}, \mathcal{X}, \mathbf{G}, t), \sigma^2_t\mathbf{I})$.
}\ENDFOR
\end{algorithmic}
\end{algorithm}

\begin{figure*}[!t]
\centering
\includegraphics[width=\textwidth]{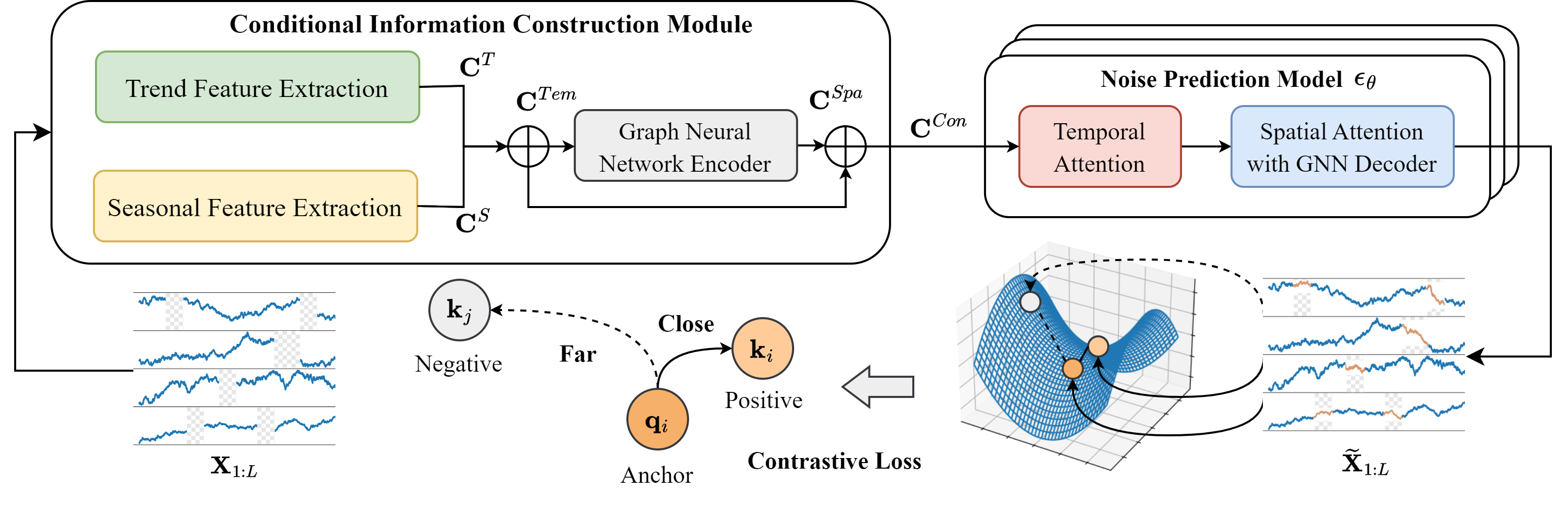}
\caption{Overview conditional information construction and noise prediction.
The Conditional Information Construction Module (Section~\ref{module:conditional}) processes observed values to learn the conditional representation $\mathbf{C}^{\mathit{Con}}$, which will later be used to guide learning spatiotemporal dependencies in the reverse process. 
The Noise Prediction Model
(Section~\ref{module:noise}) takes the noisy information $\textbf{H}^{in}$, the conditional feature $\mathbf{C}^{\mathit{Con}}$, and the adjacency matrix $\mathbf{A}$ as the input to convert noisy information into spatiotemporal data with imputed values.
Additionally, we apply contrastive learning (Section~\ref{contrastive}) to help learn discriminative features and generalize the model.
} 
\label{fig:noiseprediction}
\end{figure*}

\subsection{Conditional Information Construction Module}
\label{module:conditional}

The conditional information construction module generates a conditional representation $\mathbf{C}^{\mathit{Con}}$ to capture spatiotemporal dependencies from the observed values and the coarse interpolation, delivering input for the noise prediction model $\epsilon_\theta$.
The module works as follows:
first, we map the interpolated conditional information $\mathcal{X}$ to the latent space, $\mathbf{C}^{in} = Conv(\mathcal{X})$;
then, we use two separate modules to construct trend and seasonal representations from the intermediate representation;
finally, a Graph Neural Network Encoder aggregates the neighbor information with adjacency matrix $\mathbf{A}$.


\begin{figure}[!t]
\includegraphics[width=0.485\textwidth]{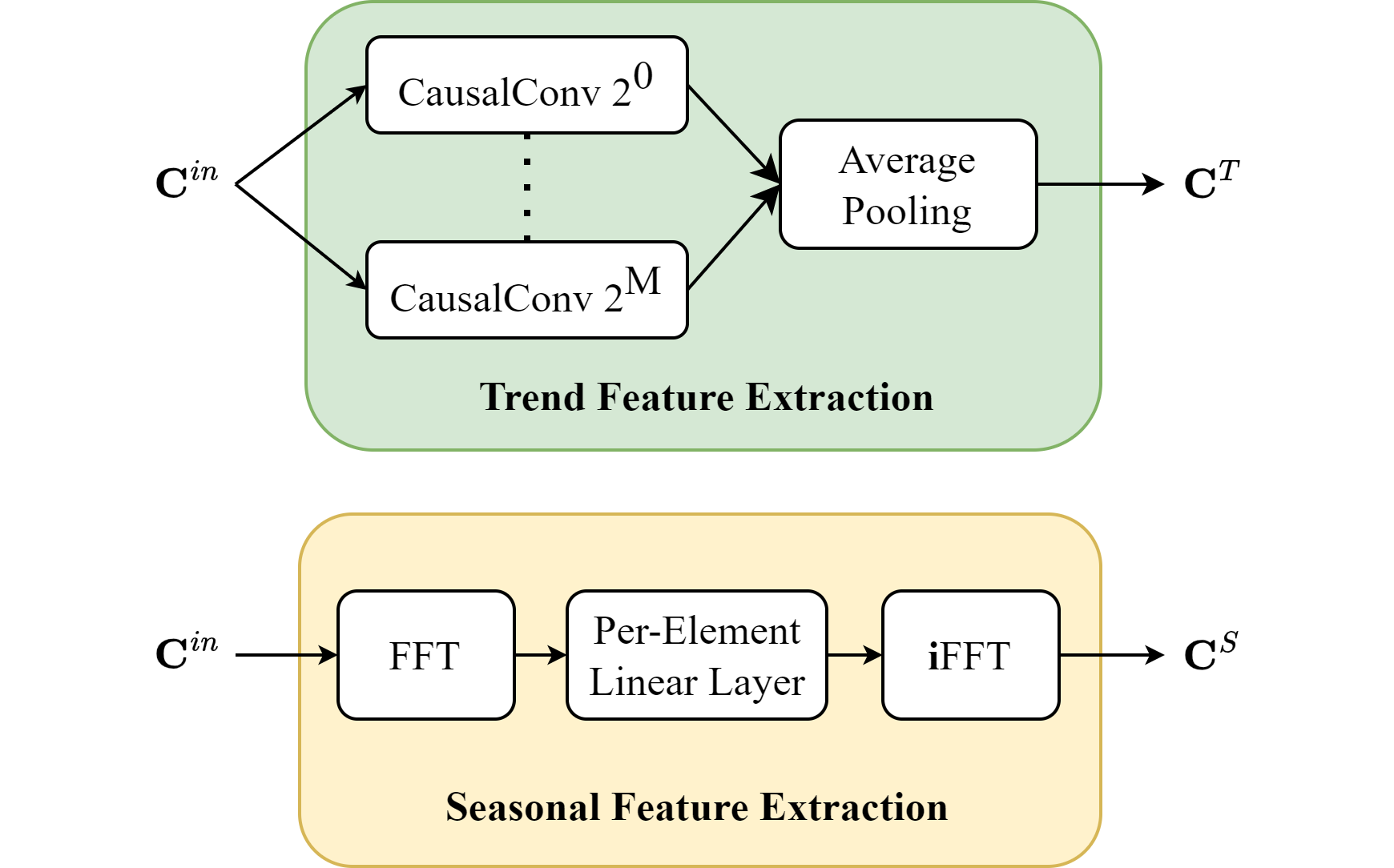}
\caption{The architecture of Trend Feature Extraction and Seasonal Feature Extraction.
The Temporal Feature Extraction (the upper box) consists of a stack of causal convolution layers with different kernel sizes. The Seasonal Feature Extraction (the lower box) is implemented by FFT and iFFT layers.} 
\label{fig:TSFD}
\end{figure}

In particular, we propose to derive separate representations for the trend and seasonal feature extraction of spatiotemporal data (shown in Fig.~\ref{fig:TSFD}, to capture complex and likely non-stationary temporal dependencies.
The Trend Feature Extraction is designed as a mixture of $M +1$ autoregressive experts, with $M = \lfloor log_2(L/2) \rfloor$. Each expert is realized through a 1-dimensional causal convolution, with $d$ input channels and $d_T$ output channels. The kernel size for the $m$-th expert is set to $2^m$. Every expert generates an output matrix $\mathbf{C}^{T,m} = CausalConv(\mathbf{C}^{in},2^m)$. To synthesize the trend representation $\mathbf{C}^T$, we apply an average-pooling operation over the collective outputs of these experts, which can be formulated as:
\begin{equation}
    \mathbf{C}^T = AvgPool(\mathbf{C}^{T,0}, \mathbf{C}^{T,1}, \cdots, \mathbf{C}^{T,M}) = \frac{1}{M+1}\sum^M_{m=0}\mathbf{C}^{T,i}.
    \label{eq:trend}
\end{equation}

We learn seasonal representations in the frequency domain, following~\cite{woo2022cost}. 
Specifically, the Seasonal Feature Extraction incorporates a learnable Fourier layer to extract seasonal features from the frequency domain. This module includes a discrete Fourier transform (DFT), which converts time domain representations into their frequency domain counterparts, denoted as $\mathcal{F}(\mathbf{C}^{in}) \in \mathbb{C}^{F \times (d \cdot N)}$. Here, $F = \lfloor L/2 \rfloor +1$ represents the number of frequencies captured. 
The subsequent learnable Fourier layer is implemented by a per-element linear layer to apply an affine transform on each frequency with different complex-valued parameters.
Following the frequency domain processing, we revert the representations to the time domain to construct the seasonal representation $\mathbf{C}^S \in \mathbb{R}^{L \times (d_S \cdot N)}$ through an inverse DFT layer. The $i$, $k$-th element of the seasonal representation can be formulated as:
\begin{equation}
    \mathbf{C}_{i,k}^S = \mathcal{F}^{-1} \left(\sum^{d}_{j=1}\mathbf{W}_{i,j,k}\mathcal{F}(\mathbf{C}_{i,j} + \mathbf{B}_{i,k})\right),
    \label{eq:season}
\end{equation}
where $\mathbf{W} \in \mathbb{C}^{F \times (d \cdot N) \times (d_S \cdot N)}$ and $\mathbf{B} \in \mathbb{C}^{F \times (d_S \cdot N)}$ represent the learnable parameters of the per-element linear layer.

The intermediate temporal representation $\mathbf{C}^{Tem}$ is the concatenation of the trend and seasonal representations, $\mathbf{C}^{Tem} = [\mathbf{C}^T;\mathbf{C}^S] \in \mathbb{R}^{L \times (d \cdot N)}$, where $\mathbf{C}^T \in \mathbb{R}^{L \times (d_T \cdot N)}$, $\mathbf{C}^S \in \mathbb{R}^{L \times (d_S \cdot N)}$, and $d = d_T + d_S$.
Besides, we adopt a GNN Encoder to capture the spatial dependencies and generate the final conditional representation, which can be formulated as:
\begin{equation}
\begin{split}
    \mathbf{C}^{Spa} &= \mathcal{G}(\mathbf{C}^{Tem},\mathbf{A}), \\
    \mathbf{C}^{\mathit{Con}} &= MLP(Norm(\mathbf{C}^{Spa} + \mathbf{C}^{Tem}),
\end{split}
\end{equation}
where $\mathbf{C}^{Spa}$ is the intermediate spatial conditional information, $\mathbf{C}^{\mathit{Con}} \in \mathbb{R}^{N \times L \times d}$ is the final conditional representation, and $\mathbf{A}$ is the adjacency matrix including the geographical relationships. 
$\mathcal{G}(\cdot)$ denotes the GNN encoder, a component that can be implemented by any GNN architecture. In our experiment, we choose the graph convolution module from Graph Wavenet~\cite{wu2019graph} to implement it. This module applies an adjacency matrix with a bidirectional distance-based matrix and an adaptive learnable matrix to capture the spatial dependencies. The conditional representation $\mathbf{C}^{\mathit{Con}}$ serves as a solution to constructing conditional information; it contains the disentangled temporal dependencies and aggregated spatial geographical relationships when compared with the initial interpolated conditional information $\mathcal{X}$. 

\subsection{Noise Prediction Model}
\label{module:noise}

The noise prediction model $\epsilon_\theta$ is used in the reverse process to sample $\widetilde{\mathbf{X}}^t$ step by step.
It uses an attention mechanism to help mitigate the randomness brought by the injected Gaussian noise to data distributions.
It also uses the conditional information $\mathbf{C}^{\mathit{Con}}$ as guidance to capture spatiotemporal dependencies in noisy data $\widetilde{\mathbf{X}}^t$.
%
The model takes three inputs: the noisy information $\mathbf{H}^{in} = Conv(\mathcal{X}||\widetilde{\mathbf{X}}^t)$, the conditional representation $\mathbf{C}^{\mathit{Con}}$, and the geographical adjacency matrix $\mathbf{A}$.
It first constructs a temporal feature $\mathbf{H}^{Tem}$ using a Temporal Attention Module $\gamma_T(\cdot)$.
Then, the output is passed to a Spatial Attention Module with GNN decoder $\gamma_S(\cdot)$ to construct a spatiotemporal feature $\mathbf{H}^{spa}$.
The above process can be formulated as:
\begin{equation}
    \begin{split}
        \mathbf{H}^{tem} = & \gamma_T(\mathbf{H}^{in}) = Attn_{tem}(\mathbf{H}^{in}), \\
        \mathbf{H}^{spa} = & \gamma_S(\mathbf{H}^{tem}, \mathbf{A}) \\
        = & MLP(Norm(Attn_{spa}(\mathbf{H}^{tem}) + \mathbf{H}^{tem}) \\
         & + Norm(\mathcal{G}(\mathbf{H}^{tem},\mathbf{A}) +\mathbf{H}^{tem})),
    \end{split}
    \label{eq:noise}
\end{equation}
where $Attn_{tem}(\cdot)$ and $Attn_{spa}(\cdot)$ are temporal and spatial attention mechanisms, respectively.

We employ the dot-product multi-head self-attention mechanism in the Transformer architecture for $Attn_{tem}(\cdot)$ and $Attn_{spa}(\cdot)$ in Eq.~\eqref{eq:noise}.
Here, we additionally use the conditional representation $\mathbf{C}^{\mathit{Con}}$ in the attention calculation to mitigate the impact of noise accumulated during diffusion steps.
Taking $Attn_{tem}(\cdot)$ for example, we formulate it as:
\begin{equation}
    \begin{split}
        Attn_{tem}(\mathcal{A}_{T},\mathcal{V}_{T}) &= SoftMax\left(\frac{\mathcal{Q}_{T}\mathcal{K}_{T}^T}{\sqrt{d}}\right) \cdot \mathcal{V}_{T},\\
    \end{split}
\end{equation}
where $W_{T}^{Q}, W_{T}^{K}, W_{T}^{V} \in \mathbb{R}^{d \times d}$ are all learnable projection parameters;
$\mathcal{Q}_{T} = \mathbf{C}^{\mathit{Con}} \cdot \mathbf{W}_{T}^{Q}$, $\mathcal{K}_{T} = \mathbf{C}^{\mathit{Con}} \cdot \mathbf{W}_{T}^{K}$, and $\mathcal{V}_{T} = \mathbf{H}^{in} \cdot \mathbf{W}_{T}^{V}$.
$Attn_{spa}(\mathcal{A}_{\mathcal{S}},\mathcal{V}_{\mathcal{S}})$ is modified in a similar way.

The noise prediction model consists of multiple stacked layers of the above attention mechanisms. The outputs $\mathbf{H}^{spa}$ from each layer are entered into residual and skip connections after a gated activation unit. The residual connection is applied as the input for the subsequent layer, while the skip connections from each layer are aggregated. This aggregated output is processed through two layers of 1-d convolution to derive the final output of the noise prediction model $\epsilon_\theta$, which only retains values of the imputation target. 

\subsection{Contrastive Learning Strategy}
\label{contrastive}
We develop a contrastive learning strategy to enhance the robustness and stability of the diffusion model.
Following~\cite{he2020momentum}, we apply two distinct residual layer modules to generate representations of the positive pair, and a dynamic dictionary with a queue to generate negative pairs. The contrastive loss is calculated as:
\begin{equation}
    \mathcal{L}_{cl}  = -\frac{1}{N} \sum_{i=1}^N \log \frac{\exp\left(\mathbf{q}_i \cdot \mathbf{k}_i/\tau\right)}{\exp\left(\mathbf{q}_i, \mathbf{k}_i/\tau\right) +\sum_{j=1}^{K} \exp\left(\mathbf{q}_i \cdot \mathbf{k}_j/\tau\right)}, \\
\label{eq:loss}
\end{equation}
where $\mathcal{L}_{cl}$ denotes the contrastive loss, which is an InfoNCE Loss widely used in unsupervised learning,
$N$ is the number of samples, $K$ is the number of negative samples, and $\tau$ is the temperature coefficient.

On the above basis, the overall model loss $\mathcal{L}$ is formulated as a combination of the reconstruction and contrastive losses:
\begin{equation}
    \mathcal{L} =  \mathcal{L}_{rl}(\theta) + \alpha \mathcal{L}_{cl},
\end{equation}
where the reconstruction loss $\mathcal{L}_{rl}(\theta)$ is computed by Eq.~\eqref{eq:losstheta}; the weight $\alpha$ determines the relative importance of the contrastive loss.

\section{Experiments}
\label{sec:experiment}
This section reports our experiments to evaluate the proposed approach, covering experimental settings (datasets, baselines, evaluation metrics), experimental design, and results analysis.

\subsection{Datasets}
We employ three datasets from the meteorology and transportation fields:
\begin{itemize}
    \item \textbf{AQI-36}~\cite{zheng2014urban}: includes hourly PM2.5 measurements recorded by 36 monitoring stations located in Beijing. The data spans a period of 12 months.
    \item \textbf{PEMS-BAY}~\cite{li2018diffusion}: includes traffic speed data collected from 325 sensors located on highways in the San Francisco Bay Area. The duration of data collection is six months.
    \item \textbf{METR-LA}~\cite{li2018diffusion}: includes traffic speed data gathered by 207 sensors located on highways in Los Angeles County. The duration of data collection is 4 months.
\end{itemize}

Each dataset in our study exhibits distinct characteristics and poses specific challenges relevant to spatiotemporal imputation. This diversity renders them suitable for evaluating the efficacy of the C$^2$TSD framework.

\subsection{Baselines} 
We evaluate the performance of C$^2$TSD vis comparisons with a selection of state-of-the-art baselines,
including statistical methods (MEAN, DA, kNN, Linear), machine learning methods (MICE, VAR, KF), matrix factorization methods (TRMF, BATF), deep learning methods (BRITS, GRIN, rGAIN), and generative methods (V-RIN, GP-VAE, CSDI, PriSTI).
The diverse baselines ensure the comparative evaluation is comprehensive.
We briefly introduce these baseline models below:

\begin{itemize}
    \item \textbf{MEAN}: This method imputes missing values using each node's historical average value.
    \item \textbf{DA}: Daily averages at relevant time steps for imputation.
    \item \textbf{kNN}: This approach calculates and uses the average value of geographically proximate nodes for imputation.
    \item \textbf{Linear}: Linear interpolation of the time series for each node, as implemented by torchcde\footnote{https://github.com/patrick-kidger/torchcde}.
    \item \textbf{KF}: A Kalman Filter is employed to impute time series data for each node, as implemented by\footnote{https://github.com/rlabbe/filterpy}.
    \item \textbf{MICE}~\cite{white2011multiple}: Multivariate Imputation by Chained Equations, based on Fully Conditional Specification, where a separate model imputes each incomplete variable.
    \item \textbf{VAR}: A vector autoregressive model used as a single-step predictor.
    \item \textbf{TRMF}~\cite{yu2016temporal}: Temporal Regularized Matrix Factorization.
    \item \textbf{BATF}~\cite{chen2019missing}: Bayesian Augmented Tensor Factorization integrates spatiotemporal domain knowledge.
    \item \textbf{BRITS}~\cite{cao2018brits}: A bidirectional RNN-based method for multivariate time series imputation.
    \item \textbf{GRIN}~\cite{cini2021filling}: This model uses a bidirectional GRU combined with a graph convolution network for multivariate time series imputation.
    \item \textbf{rGAIN}~\cite{yoon2018gain}: An approach that extends the GAIN model with a bidirectional recurrent encoder-decoder structure.
    \item \textbf{V-RIN}~\cite{mulyadi2021uncertainty}: A method to improve deterministic imputation using VAE. The quantified uncertainty provides the probability imputation result.    
    \item \textbf{GP-VAE}~\cite{fortuin2020gp}: A probabilistic imputation method for time series that combines a Variational Autoencoder with Gaussian processes.
    \item \textbf{CSDI}~\cite{tashiro2021csdi}: A probability imputation method based on a conditional diffusion probability model, which treats different nodes as multiple feature dependencies.
    \item \textbf{PriSTI}~\cite{liu2023pristi}: A conditional diffusion framework with a feature extraction module to calculate the spatiotemporal attention weights.
\end{itemize}


\subsection{Evaluation Metrics}
Following~\cite{liu2023pristi}, we apply three metrics for performance evaluation: \textit{Mean Absolute Error (MAE)} and \textit{Mean Squared Error (MSE)} offer a quantitative assessment of the error between the imputed values and the ground truth;
\textit{Continuous Ranked Probability Score (CRPS)}~\cite{hersbach2000decomposition}
calculates the similarity between the estimated probability distribution and its observed values.
It measures a model's robustness in harnessing the uncertainty in predictions and is particularly useful for evaluating the quality of probabilistic forecasts.

Given a missing value $x$ with its estimated probability distribution $D$, CRPS quantifies the compatibility of $x$ with $D$. This compatibility is determined through the integral of the quantile loss $\Lambda_\alpha$, which is defined as:
\begin{equation}
    \begin{split}
        \text{CRPS}(D^{-1}, x) &= \int^1_0 2\Lambda_\alpha(D^{-1}(\alpha),x)d\alpha, \\
        \Lambda_\alpha(D^{-1}(\alpha),x) &= (\alpha - \mathbb{1}_{x<D^{-1}(\alpha)})(x-D^{-1}(\alpha)),
    \end{split}
\end{equation}
where the parameter $\alpha \in [0,1]$ represents the quantile levels, while $D^{-1}(\alpha)$ denotes the $\alpha$-quantile of the distribution $D$, and $\mathbb{1}$ serves as the indicator function. Consistent with~\cite{liu2023pristi}, we generate 100 samples to accurately simulate our model's distribution of missing values. The quantile losses are calculated for discretized quantile levels using intervals of 0.05, resulting in 19 evenly spaced points within the range of $(0, 1)$. This calculation process is formulated as:
\begin{equation}
    \text{CRPS}(D^{-1},x) \simeq \sum^{19}_{i=1} 2\Lambda_{i \times 0.05} (D^{-1}(i \times 0.05), x)/19.
\end{equation}

We calculate CRPS for each estimated missing value and use the average CRPS to evaluate our model's probabilistic forecasting accuracy, as formulated as:
\begin{equation}
    \text{CRPS}(D,\widetilde{X}) = \frac{\sum_{\tilde{x} \in \widetilde{X}} \text{CRPS}(D^{-1},\tilde{x})}{|\widetilde{X}|}.
\end{equation} 

\subsection{Experimental Design}
\label{sec:settings}
\subsubsection{Datasets Split}
We 
divided the data into training, validation, and test sets in alignment with the dataset division methodology employed by~\cite{cini2021filling}.
For the AQI-36 dataset, March, June, September, and December were selected
as the test set. The validation set includes the last 10\% of the data from February, May, August, and November, while the remaining forms the training set.
As for the PEMS-BAY and METR-LA datasets, a chronological division strategy was applied, with 70\% of the data allocated to the training set, 10\% to the validation set, and the rest to the test dataset. 

\subsubsection{Choice of Imputation Targets}
Although the datasets inherently contain missing values, they lack the ground truth for evaluation purposes.
Therefore, the common practice in the field of research is to artificially introduce missing values into test portions of each dataset to facilitate the performance evaluation.
For the AQI-36 dataset, we simulated the distribution of real-world missing data, creating a relevant setting for evaluating imputation performance following~\cite{yi2016st}.
For the traffic datasets: EMS-BAY and METR-LA, we used a mask matrix~\cite{cini2021filling} to manually inject missing values.
Specifically, we employed two missing patterns to create the masks for missing values: (1) Point missing, where we randomly masked 25\% of the observations, and (2) Block missing, which randomly masks 5\% of the observations and then further masking consecutive observations ranging from 1 to 4 hours for each sensor with a probability of 0.15\%.

\subsubsection{Training Environment and Hyperparameters}
We trained our model on an Nvidia A40 GPU equipped with 48 GB memory and applied the Adam optimizer, well-known for its adaptability and robust performance
to optimize C$^2$TSD's parameters.
Additionally, we applied a cosine annealing decay strategy to the learning rate. This strategy gradually reduced the learning rate following a cosine function to fine-tune the model's performance over time.
%
We configured hyperparameters to balance the learning efficiency and performance.
Specifically, we set the learning rate to $10^{-3}$ and the decay to $10^{-5}$ to enable effective learning while preventing overshooting during optimization.
Regarding the diffusion process (Eq.~\eqref{eq1}), the diffusion step $T$ was set to 100 for the AQI-36 dataset and 50 for two traffic datasets; the minimum ($\beta_1$) and maximum ($\beta_T$) noise levels were set to $10^{-4}$ and $0.2$, respectively.



\begin{table*}[t]
\renewcommand{\arraystretch}{1.3}
  \centering
  \caption{Peformance of compared methods with respect to MAE and MSE, The best result is boldfaced. The second-best result is underlined. \textit{Improv.} denotes the performence improvement of C$^2$TSD over the best baseline.}
  \renewcommand{\arraystretch}{1.15}{
    \begin{tabular}{lrrrrrrrrrr}
    \hline
    \multicolumn{1}{c}{\multirow{3}{*}{Method}} & \multicolumn{2}{c}{\multirow{2}{*}{AQI-36}} & \multicolumn{4}{c}{PEMS-BAY} & \multicolumn{4}{c}{METR-LA} \\
\cline{4-11}          &       &       & \multicolumn{2}{c}{Point missing} & \multicolumn{2}{c}{Block missing} & \multicolumn{2}{c}{Point missing} & \multicolumn{2}{c}{Block missing} \\
\cline{2-11}          & \multicolumn{1}{r}{MAE} & \multicolumn{1}{c}{MSE} & \multicolumn{1}{r}{MAE} & \multicolumn{1}{c}{MSE} & \multicolumn{1}{r}{MAE} & \multicolumn{1}{c}{MSE} & \multicolumn{1}{r}{MAE} & \multicolumn{1}{c}{MSE} & \multicolumn{1}{r}{MAE} & \multicolumn{1}{c}{MSE} \\
    \hline
    Mean  & 53.48 & 4578.08 & 5.42  & 86.59 & 5.46  & 87.56 & 7.56  & 142.22 & 7.48  & 139.54 \\
    DA    & 50.51 & 4416.10 & 3.35  & 44.50 & 3.30  & 43.76 & 14.57 & 448.66 & 14.53 & 445.08 \\
    kNN   & 30.21 & 2892.31 & 4.30  & 49.80 & 4.30  & 49.90 & 7.88  & 129.29 & 7.79  & 124.61 \\
    Linear & 14.46 & 673.92 & 0.76  & 1.74  & 1.54  & 14.14 & 2.43  & 14.75 & 3.26  & 33.76 \\
    \hline
    KF    & 54.09 & 4942.26 & 5.68  & 93.32 & 5.64  & 93.19 & 16.66 & 529.96 & 16.75 & 534.69 \\
    MICE  & 30.37 & 2594.06 & 3.09  & 31.43 & 2.94  & 28.28 & 4.42  & 55.07 & 4.22  & 51.07 \\
    VAR   & 15.64 & 833.46 & 1.30  & 6.52  & 2.09  & 16.06 & 2.69  & 21.10 & 3.11  & 28.00 \\
    TRMF  & 15.46 & 1379.05 & 1.85  & 10.03 & 1.95  & 11.21 & 2.86  & 20.39 & 2.96  & 22.65 \\
    BATF  & 15.21 & 662.87 & 2.05  & 14.90 & 2.05  & 14.48 & 3.58  & 36.05 & 3.56  & 35.39 \\
    \hline
    BRITS  & 14.50 & 622.36 & 1.47  & 7.94  & 1.70  & 10.50 & 2.34  & 16.46 & 2.34  & 17.00 \\
    GRIN  & 12.08 & 523.14 & 0.67  & 1.55  & 1.14  & 6.60  & 1.91  & 10.41 & 2.03  & 13.26 \\
    V-RIN & 10.00 & 838.05 & 1.21  & 6.08  & 2.49  & 36.12 & 3.96  & 49.98 & 6.84  & 150.08 \\
    GP-VAE & 25.71 & 2589.53 & 3.41  & 38.95 & 2.86  & 26.80 & 6.57  & 127.26 & 6.55  & 122.33 \\
    rGAIN & 15.37 & 641.92 & 1.88  & 10.37 & 2.18  & 13.96 & 2.83  & 20.03 & 2.90  & 21.67 \\
    CSDI  & 9.51  & 352.46 & \textbf{0.57} & \underline{1.12}  & 0.86  & 4.39  & 1.79  & 8.96  & 1.98  & 12.62 \\
    PriSTI & \underline{9.22}  & \underline{324.95} & \textbf{0.57} & \textbf{1.09} & \textbf{0.78} & \underline{3.50} & \underline{1.76}  & \underline{8.63}  & \underline{1.89}  & \underline{11.21} \\
    \hline
    C$^2$TSD & \textbf{9.09} & \textbf{309.81} & \textbf{0.57} & \textbf{1.09} & \underline{0.81}  & \textbf{3.23}  & \textbf{1.70} & \textbf{8.47} & \textbf{1.79} & \textbf{10.65} \\
    \textit{Improv.} & \textit{1.43\%} & \textit{4.89\%} & \textit{-} & \textit{-} & \textit{-} & \textit{8.48\%} & \textit{3.28\%} & \textit{1.95\%} & \textit{5.91\%} & \textit{5.24\%}  \\
    \hline
    \end{tabular}}
  \label{tab:main}%
\end{table*}%

\begin{table}
\renewcommand{\arraystretch}{1.3}
    \centering
    \caption{Performance of generative methods in CRPS. 'Point' and 'Block' denote the respective missing patterns.}
    \renewcommand{\arraystretch}{1.15}{
    \begin{tabular}{lrrrrr}
    \hline
    \multicolumn{1}{c}{\multirow{2}{*}{Method}} & \multicolumn{1}{c}{\multirow{2}{*}{AQI-36}} & \multicolumn{2}{c}{PEMS-BAY} & \multicolumn{2}{c}{METR-LA} \\
          &       & \multicolumn{1}{c}{Point} & \multicolumn{1}{c}{Block} & \multicolumn{1}{c}{Point} & \multicolumn{1}{c}{Block} \\
    \hline
    V-RIN & 0.3154 & 0.0191 & 0.3940 & 0.0781 & 0.1283 \\
    GP-VAE & 0.3377 & 0.0568 & 0.0436 & 0.0977 & 0.1118 \\
    CSDI  & 0.1056 & \textbf{0.0067} & 0.0127 & 0.0235 & 0.0260 \\
    PriSTI & \underline{0.1025} & \textbf{0.0067} & \textbf{0.0094} & \underline{0.0231} & \underline{0.0249} \\
    \hline
    C$^2$TSD  & \textbf{0.0995} & \textbf{0.0067} & \underline{0.0097} & \textbf{0.0223} & \textbf{0.0237} \\
    \hline
    
    \end{tabular}}
    \label{tab:crps}%
    
\end{table}%

\subsection{Overall Performance}
\label{comparision}
We compare the performance of C$^2$TSD against baseline methods in terms of MAE and MSE in Table~\ref{tab:main}.
For methods that can produce probability distributions of missing values, e.g., V-RIN, GP-VAE, CSDI, PriSTI, and C$^2$TSD, we additionally show their comparisons in CRPS in Table~\ref{tab:crps}.
Specifically, given a missing value, we randomly generated 100 samples to simulate a probability distribution; then, the deterministic imputation result was determined by taking the median of these samples.

The results in Table~\ref{tab:main} and Table~\ref{tab:crps} show our proposed C$^2$TSD consistently outperformed the compared baselines across the experimental datasets and missing patterns, while PriSTI achieved the second-best performance in most cases.
%
More complex methods (e.g., those based on RNNs and generative models) are generally more effective in capturing complex temporal and spatial dependencies in real-world scenarios, thus beating statistical, machine learning, and matrix factorization-based methods, reflected by results in Table~\ref{tab:main}.
Statistical and machine learning-based methods generally rely on strict assumptions about data, e.g., stationary in time series, to predict missing values; these assumptions, however, may not align with or fully respect the complex temporal and spatial dependencies in real-world scenarios. Matrix factorization-based methods, similarly, rely on the low-rank assumption, which may not perfectly match the real data, resulting in their inferior performance.

GRIN outperformed RNN-based methods (rGAIN and BRITS) as it additionally leveraged spatial dependencies through graph convolution networks.
However, it still fell behind diffusion-based methods (e.g., CSDI and PriSTI) due to the inherent limitation of error accumulation in recurrent structure models
Among deep generative models, diffusion-based methods (CSDI, PriSTI, and our C$^2$TSD) consistently beat VAE-based methods (V-RIN and GP-VAE).
Our method, C$^2$TSD, surpassed PriSTI in almost all metrics across the datasets and missing patterns---it ranked second only in two cases.
This result validates the overall effectiveness of our proposed approach
in spatiotemporal imputation.

\subsection{Ablation Study}

    

\begin{table}[t]
    \centering
    \renewcommand{\arraystretch}{1}
    \caption{Ablation Studies. 'Point' and 'Block' denote the respective missing patterns.}
    \renewcommand{\arraystretch}{1.15}{
    \begin{tabular}{lrrrrrr}
    \hline
    \multicolumn{1}{c}{\multirow{3}{*}{Method}} & \multicolumn{2}{c}{\multirow{2}{*}{AQI-36}} & \multicolumn{4}{c}{METR-LA} \\
    & \multicolumn{2}{c}{} & \multicolumn{2}{c}{Point} & \multicolumn{2}{c}{Block} \\
    \cline{2-7} & \multicolumn{1}{c}{MAE} & \multicolumn{1}{c}{MSE} & \multicolumn{1}{c}{MAE} & \multicolumn{1}{c}{MSE} & \multicolumn{1}{c}{MAE} & \multicolumn{1}{c}{MSE} \\
    \hline
    C$^2$TSD  & \textbf{9.09} & \textbf{309.81} & \textbf{1.70} & \textbf{8.47} & \textbf{1.79} & \textbf{10.65} \\
    \textit{w/o CL} & 10.18 & 389.23 & 2.18  & 16.83 & 2.27  & 17.83 \\
    \textit{w/o TFD} & 10.06 & 380.94 & \underline{2.07}  & \underline{13.66} & 2.14  & 15.13 \\
    \textit{w/o SFD} & \underline{9.80}  & \underline{361.93} & 2.09  & 14.70 & \underline{2.07}  & \underline{14.19} \\
    \hline
    \end{tabular}}
    \label{tab:ablation}%
\end{table}%

We evaluate the impact of key components of our approach
by comparing it with the following variants:

\begin{itemize}
    \item \textit{w/o CL}: Remove the contrastive loss $\mathcal{L}_{cl}$ in Eq.~\eqref{eq:loss}; only calculate the gradient of the reconstruction loss $\mathcal{L}_{rl}$;
    \item \textit{w/o TFD}: Remove the trend representation $\mathbf{C}^{T}$ in Eq.~\eqref{eq:trend} in the conditional information construction module; only keep the seasonal representation $\mathbf{C}^{S}$ for using to construct the conditional information;
    \item \textit{w/o SFD}: Opposite to the previous variant, remove the seasonal representation $\mathbf{C}^{S}$ in Eq.~\eqref{eq:season} in the conditional information construction module; only keep the trend representation $\mathbf{C}^{T}$ for the conditional information. 
\end{itemize}

The results (Table~\ref{tab:ablation}) show each individual component (contrastive loss, trend representation, seasonal representation) contributes to the superior performance of C$^2$TSD.
We omit to show the results on the PEMS-BAY dataset, which are consistent with our observations in Table~\ref{tab:ablation}.
Among the three components studied, the contrastive learning strategy has the greatest influence on C$^2$TSD's performance---removing it results in a dramatic drop in performance.
The trend component generally contributes more than the seasonal component based on a performance comparison between \textit{w/o Trend} and \textit{w/o Season}.
Removing the trend component caused a bigger performance drop on the AQI-36 dataset under both missing patterns and the METR-LA dataset under the block missing pattern, according to Table~\ref{tab:ablation}.
The results on the AQI-36 dataset can be explained by the complex factors that influence PM2.5 concentration and the lack of clear periodicity on the dataset.
The seasonal component is more important than the trend component on the METR-LA dataset under the point missing pattern. A possible reason is that the average traffic speed typically does not exhibit rapid trend drifts within short intervals, which are more common in the point-missing pattern.

\subsection{Hyperparameter Analysis}


\begin{figure}[!t]
\includegraphics[width=0.485\textwidth]{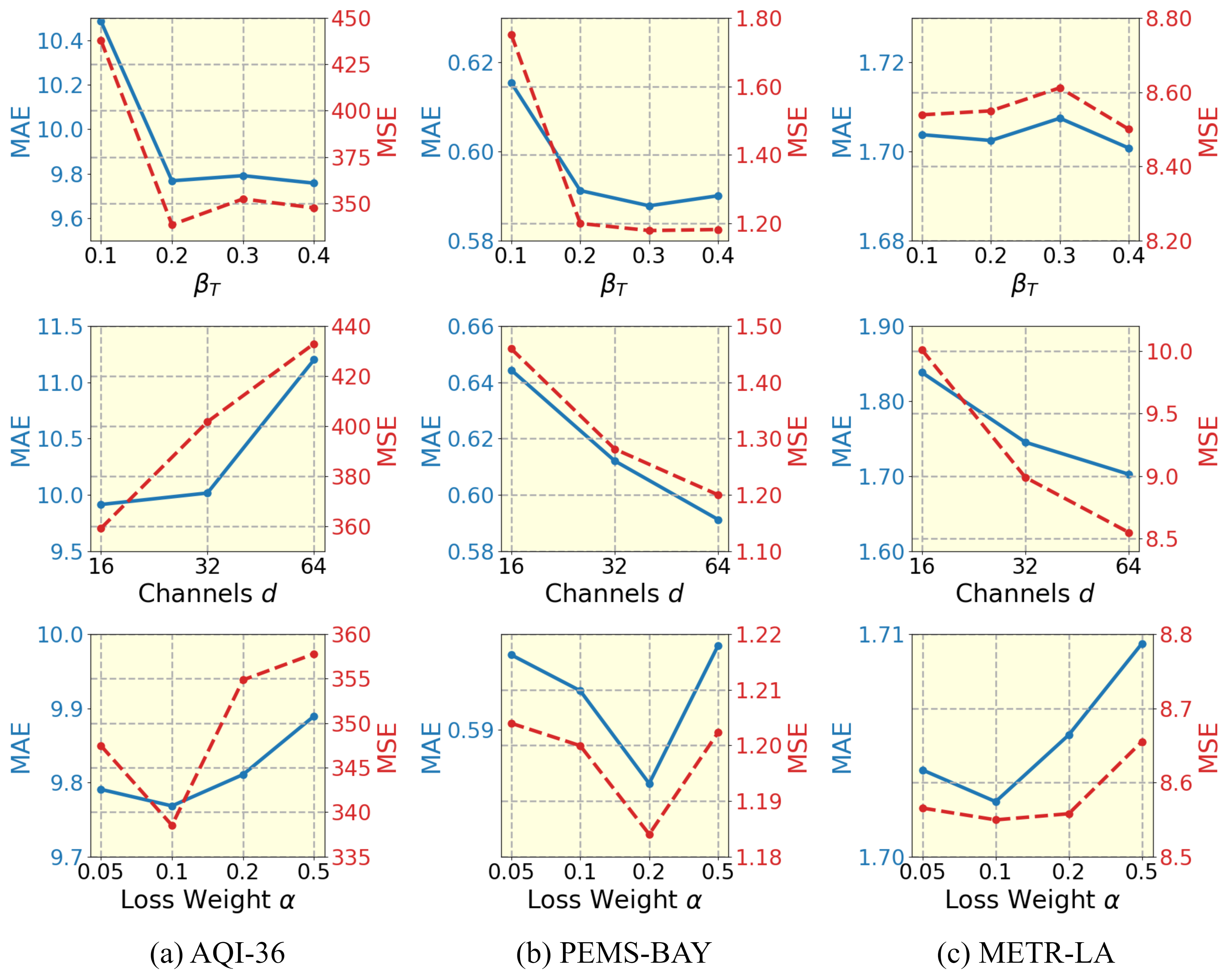}
\caption{Sensitivity study of key hyperparameters.} 
\label{fig:analysis}
\end{figure}

\subsubsection{Maximum noise level, $\beta_T$}
This parameter regulates the noise intensity during the diffusion steps.
Our results (Fig.~\ref{fig:analysis}) show our model sustains relatively low MAE and MSE when $\beta_T$ ranges from 0.2 to 0.4 for the AQI-36 and PEMS-BAY datasets.
A smaller $\beta_T$ value, e.g., 0.1, might result in a slow diffusion process, requiring more iterations for the model to effectively convert noisy data to the target data distribution. 
For the METR-LA dataset, the model is less impacted by $\beta_T$ values, showing stable performance in MAE and MSE.
We could not experiment on cases where $\beta_T$ was larger than 0.4, limited by the GPU.
We have set $\beta_T$ to 0.2 during our comparison experiments (Section~\ref{comparision}) to ensure optimal performance.

\subsubsection{Channel size of the hidden state, $d$}
The channel size $d$ plays a pivotal role in defining the model's capacity to encode information, and the optimal channel size differs across the datasets, according to our results (Fig.~\ref{fig:analysis}), i.e., 16 for the AQI-36 dataset and 64 for the PEMS-BAY and METR-LA datasets.
The AQI-36 dataset has fewer variables than the traffic datasets; thus, a smaller channel size would suffice for the model to learn spatiotemporal embeddings. A smaller size prevents the model from incorporating unnecessary information into the embeddings and thus impairs performance.
A smaller channel size than 64 may not empower the model with sufficient capacity to capture complex spatiotemporal dependencies for the other two datasets.

\subsubsection{Weight of the contrastive loss, $\alpha$}
The hyperparameter $\alpha$ balances the weights of the reconstruction loss and the contrastive loss.
It should be customized for specific datasets to avoid overlooking either of the losses.
Our results (Fig.~\ref{fig:analysis}) show C$^2$TSD performed best when $\alpha$ was set to 0.1 for the AQI-36 and METR-LA datasets and 0.2 for the PEMS-BAY dataset.
Either a larger value or a smaller value would lead to sub-optimal performance. 
Compared to the maximum noise level $\beta_T$ and channel size $d$, the model performance maintains more stability when setting the contrastive loss weight $\alpha$ in a range of $[0.05,0.5]$.

\section{Related Work}
\label{sec:related}
\subsection{Traditional Methods for Time Series Imputation}

A range of statistical and machine-learning methods have been employed to estimate the missing values, e.g., autoregressive moving average (ARMA)~\cite{ansley1984estimation}, expectation-maximization (EM)~\cite{shumway1982approach}, and $k$-nearest neighbors (kNN)~\cite{hastie2009elements}. These models, however, typically rely on strong assumptions about the \textit{temporal stationarity} and \textit{inter-series similarity} of the times series data to make estimations. Those assumptions do not necessarily capture the complexities of real-world multivariate time series data, leading to potentially unsatisfactory performance in practical scenarios.
Moreover, low-rank matrix factorization has emerged as another viable method for spatiotemporal imputation. This technique leverages intrinsic spatial and temporal patterns informed by prior knowledge. Notable implementations include TRMF~\cite{yu2016temporal} and BATF~\cite{chen2019missing}. The former integrates temporal dependencies into a temporal regularized matrix factorization framework; the latter infuses domain knowledge from transportation systems into an augmented tensor factorization model for traffic data. Additionally, TIDER~\cite{liu2022multivariate} combines matrix factorization with multivariate time series disentanglement representation.

Deep learning-based methods, especially RNNs, GANs and their variants, are increasingly applied to multivariate time series imputation. GRU-D~\cite{che2018recurrent} integrates masking and time interval representations into its deep model architecture, using recurrent structures to capture long-term temporal relationships. Subsequently, BRITS~\cite{cao2018brits} introduced a bidirectional RNN-based model capable of predicting multiple correlated missing values in time series. This model handles extensive correlated missing values and adapts to nonlinear dynamics, supporting a data-driven imputation approach.
GAN-based methods have also shown considerable efficacy in imputing missing values. These methods enable the generator to learn the entire data distribution, predicting missing values through iterative adversarial processes against the discriminator~\cite{yoon2018gain,luo2019e2gan}. 
Graph neural networks (GNNs)-based approaches have been employed to facilitate learning spatial dependencies in addressing the imputation task. For instance, STGNN-DAE~\cite{kuppannagari2021spatio} uses power grid topology and time series data from each grid meter, accounting for both spatial and temporal correlations. Another notable work, GRIN~\cite{cini2021filling}, introduces a spatial-temporal encoder that amalgamates variable and temporal dependencies.
While showing strengths in capturing spatiotemporal dependencies,
recurrent methods cannot avoid the error accumulation problem, while graph-based models exhibit lower robustness to noisy, incomplete, or unfamiliar data, as their deterministic output mechanism may not effectively adapt to non-stationary input data.

\subsection{Diffusion Method in Multivariate Time Series}

Diffusion models have been widely applied in forecasting, anomaly detection, and imputation tasks on multivariate time series data.
DDPMs~\cite{ho2020denoising} represent the first diffusion models applied to forecasting tasks.
TimeGrad~\cite{rasul2021autoregressive} injects noise into data at each predictive time point, subsequently employing a backward transition kernel conditioned on historical time series for gradual denoising; it uses the hidden state calculated with a recurrent structure to simulate the conditional distribution.
ScoreGrad~\cite{yan2021scoregrad} shares the same target distribution as TimeGrad but uses a conditional SDE-based score-matching module, converting the diffusion process from discrete to continuous and substituting the number of diffusion steps with an integration interval.
D$^3$VAE~\cite{li2022generative} first focuses on dealing with limited and noisy time series data using a bidirectional variational autoencoder (BVAE). 
Recent work has introduced graph structure to times series tasks. 
DiffSTG~\cite{wen2023diffstg} proposes a UGnet, a combination of a U-Net-based network and GNN, to process spatiotemporal dependencies.
GCRDD~\cite{li2023graph} designs a graph-modified GRU, which calculates the weight matrix in GRU with a graph convolution module in the noise-matching network. 

Diffusion models are also applied to time series imputation, where spatiotemporal imputation is a special type.
CSDI~\cite{tashiro2021csdi} is a seminal DDPM-based work on diffusion-based time series imputation, which adopts the DiffWave as the noise estimation network. 
SSSD~\cite{alcaraz2022diffusion} replaces the noise estimation network with a structured state space model to capture long-term time series dependencies.
PriSTI~\cite{liu2023pristi} deals with the imputation problem on spatiotemporal data and first considers the geographic relationship for learning spatial information among different sensors.
Despite the above efforts to learn spatial and temporal features, no existing study has sought to disentangle temporal representations, limiting their capability to capture complex and non-stationary temporal dependencies in real-world scenarios.

\section{Conclusion}
\label{sec:conclusion}
We introduce C$^2$TSD, a
conditional diffusion framework that integrates temporal disentanglement and contrastive learning for spatiotemporal imputation. 
Our model predicts missing values by employing a reverse process enhanced by contrastive learning and disentangled temporal (trend and seasonality) dependencies as key conditional information to guide this process.
Our framework distinguishes itself with a unique reverse process that is guided by the disentanglement of temporal elements—trend and seasonality—as essential conditional representations. Besides, the contrastive learning strategy also helps learning spatiotemporal dependencies and enhances the stability and generalization of C$^2$TSD. 
Our model has demonstrated superior
performance to a number of state-of-the-art baselines on various datasets under different missing patterns of spatiotemporal data.


For future work,
we will consider leveraging prior knowledge of spatiotemporal dependencies to design conditional information.
Given that choosing and designing the noise prediction model is still a handcrafted task, it is worthwhile to comprehensively evaluate the existing options to assist with the decisions.
A limitation of most existing diffusion models is the high computational cost, so simplifying diffusion models for large-scale scenarios might be a promising direction.


\bibliographystyle{IEEEtran}
\bibliography{main}

\end{document}